\documentclass{scrartcl} 
\usepackage{authblk}

\usepackage{hyperref}
\usepackage{float}
\floatstyle{plaintop}
\restylefloat{table}
\usepackage{graphicx}
\usepackage{multirow}
\usepackage[labelfont=bf, ]{caption}

\begin{document}
\title{Dataset on Bi- and Multi-Nucleated Tumor Cells in Canine Cutaneous Mast Cell Tumors}

\author[1]{Christof~A.~Bertram}
\author[2]{Taryn~A.~Donovan}
\author[3]{Marco~Tecilla}
\author[1]{Florian~Bartenschlager}
\author[1]{Marco~Fragoso}
\author[1]{Frauke~Wilm}
\author[4]{Christian~Marzahl}
\author[1]{Katharina~Breininger}
\author[4]{Andreas~Maier}
\author[1]{Robert~Klopfleisch}
\author[5,4]{Marc Aubreville}

\affil[1]{Institute of Veterinary Pathology, Freie Universit\"at Berlin, Berlin, Germany}
\affil[2]{Department of Anatomic Pathology, Animal Medical Center, New York, USA}
\affil[3]{Roche Pharma Research and Early Development (pRED), Pharmaceutical Sciences, BIOmics and Pathology - Roche Innovation Center Basel, Switzerland}
\affil[4]{Pattern Recognition Lab, Computer Sciences, Friedrich-Alexander-Universit\"at Erlangen-N\"urnberg, Erlangen, Germany}
\affil[5]{Technische Hochschule Ingolstadt, Ingolstadt, Germany}
\date{}

\maketitle
\vspace{-5em}
\begin{center}
\url{christof.bertram@fu-berlin.de}
\end{center}

\begin{abstract}
Tumor cells with two nuclei (binucleated cells, BiNC) or more nuclei (multinucleated cells, MuNC) indicate an increased amount of cellular genetic material which is thought to facilitate oncogenesis, tumor progression and treatment resistance. In canine cutaneous mast cell tumors (ccMCT), binucleation and multinucleation are parameters used in cytologic and histologic  grading schemes (respectively) which correlate with poor patient outcome. For this study, we created the first open source data-set with 19,983 annotations of BiNC and 1,416 annotations of MuNC in 32 histological whole slide images of ccMCT. Labels were created by a pathologist and an algorithmic-aided labeling approach with expert review of each generated candidate. A state-of-the-art deep learning-based model yielded an $F_1$ score of 0.675 for BiNC and 0.623 for MuNC on 11 test whole slide images. In regions of interest ($2.37 mm^2$) extracted from these test images, 6 pathologists had an object detection performance between 0.270 - 0.526 for BiNC and 0.316 - 0.622 for MuNC, while our model archived an $F_1$ score of 0.667 for BiNC and 0.685 for MuNC. This open dataset can facilitate development of automated image analysis for this task and may thereby help to promote standardization of this facet of histologic tumor prognostication.  
\end{abstract}

\section{Introduction}
Microscopic evaluation of tumor biopsies can yield important information pertaining to the biological behaviour of a tumor obtained from a patient. Depending upon the tumor type, different microscopic characteristics  are combined to grading schemes, which are useful estimators of patient outcome. For canine cutaneous mast cell tumors (ccMCT), a frequent skin tumor of dogs, the current grading system encompasses counting the number of mitosis (cells undergoing division), number of multinucleated cells (MuNC) and cells with aberrant nuclear size and shape in an tumor area of $2.37 mm^2$ \cite{04}. As opposed to a single nucleus in most mast cells, MuNC contain three or more nuclei. Tumor cells with two nuclei (binucleated cells, BiNC) have not been evaluated as an prognostic parameter in histologic sections in previous studies, however, studies on cytologic specimens of ccMCT revealed a negative correlation to patient outcome \cite{07}. 

Formation of BiNC and MuNC results in increased numbers of chromosomes (genetic material) per cell (polyploidy). This augments the metabolic capacities of the cell, which is an effective strategy in coping with escalating requirements for tumor growth. Polyploidy is considered to be key actuator of oncogenesis, tumor progression, and chemotherapy resistance \cite{08,09}. Additional nuclei can be acquired by 1) fusion with other neoplastic or non-neoplastic cells (syncytia) or 2) by an incomplete cell cycle (endoreplication) in the absence of cell division (failure of cytokinesis). During normal mitosis, the chromosomes are duplicated and divided into two nuclei, which are further separated into two daughter cells. If this last step is aborted then both nuclei will remain in the cell of origin.

Deep learning-based algorithms are considered a powerful tool for reproducible automated image analysis (for example for mitotic figures), however, they require large amounts of labeled data for training and testing models \cite{02,03}. In the present work we present the first open dataset on BiNC and MuNC in ccMCT, establish a baseline performance for deep-learning based pipeline and compare algorithmic performance to six veterinary pathologists. 

\section{Materials and methods}
For this study we developed a novel set of labels for 32 publicly available whole slides images (WSI, resolution of 0.25 microns per pixel) of ccMCT. WSI were initially provided by Bertram et al. \cite{03} for a mitotic figure dataset, which included 44,880 mitotic figure labels, on the same images.

\subsection{Labeling of bi- and multi-nucleated tumor cells}
One pathologist screened the 32 WSI using the annotation software SlideRunner \cite{01} and labeled BiNC and MuNC as separate label classes. Thereby, 10,381 labels of BiNC and 775 labels of MuNC were obtained. 
Because omission of target cells during manual screening was considered a major limitation, we decided to additionally use an algorithmic-aided pipeline to identify potential target cells. Each of the candidates was subsequently reviewed by a pathologist. Using the labeling protocol described in Bertram et al. \cite{03}, we split the manual database into three folds and used each fold to train a deep learning-based network (as described below in section 2.2) in order to detect overlooked candidates. A high sensitivity of finding additional BiNC and MuNC candidates was reached by using a low detection cutoff, which intentionally resulted in in many false positive detections in order to reduce a confirmation bias of the reviewing pathologist. 
A total of 66,585 potential BiNC and 6,958 MuNC candidates with model scores between 0.3 and 1.0 were retrieved and extracted as 128$\times$128\,px images for expert review. Of these, 9,602 (14.4\,\%) were classified as BiNC and 641 (9.2\,\%) as MuNC, which increased the label classes by 92.5\,\% and 82.7\,\%, respectively, compared to the manual database. Patches that were not assigned to these two classes were useful as hard negatives in training a classification network (see below). The algorithmic-augmented database is freely accessible as a SQLite3 document via  Github. For each annotation, this database provides center coordinates (x,y) on the WSI
The final algorithmic-augmented datasets includes 19,983 labels of BiNC and 1,416 labels of MuNC. 
All code and the labels can be accessed on our github project page: https://github.com/DeepPathology/CCMCT\_Bi\_Multinucleated.

\begin{figure}[htb]
\includegraphics[width=\textwidth]{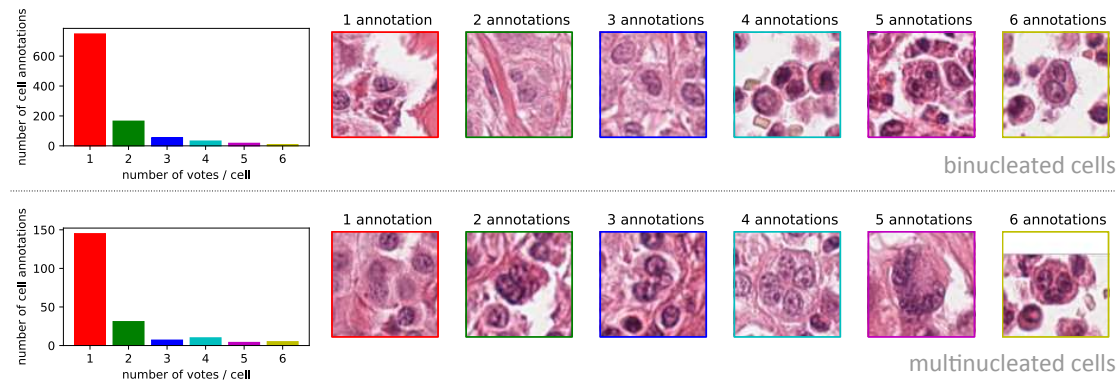}
\caption{Number of pathologists agreeing upon an object being a binucleated (upper images) and multinucleated (lower images) tumor cell.}
\label{fig1}
\end{figure}

\subsection{Dataset validation}
In order to establish a baseline for automatic detection, we trained a customized deep learning model. WSI were split into training (N = 21) and test cases (N = 11) according to the original publication of the images \cite{03}. 
Our model consists of two stages as previously described for automated mitotic figure detection \cite{02}. The primary object detection stage is a customized 
RetinaNet \cite{05} 
with a pre-trained ResNet-18 
stem \cite{06}. The second stage is a patch classifier, which was also derived from a pre-trained ResNet-18 architecture \cite{06},
and was used to differentiate hard negatives with high visual similarity to BiNC and MuNC.

\subsubsection{Performance validation on the ground truth dataset.}
First, our model was validated against the final dataset of the test images. This task evaluated the object detection performance as per the $F_1$ score on entire histologic sections and can serve as a baseline performance for future research projects. 

\subsubsection{Performance validation against pathologists.}
In order to evaluate whether algorithms can approximate the performance of pathologists, we compared object detections from both on smaller tumor areas (region of interest with a size of $2.37 mm^2$) from the 11 test WSI. Regions of interest were extracted from the WSI with the following criteria for the tumor region: images containing the most MuNC (as per dataset ground truth), or, if no MuNC were present in the WSI, images containing the most BiNC. For calculation of performance, ground truth (GT) labels comprised of the algorithmically augmented database (dataset ground truth), agreement on labels by 2/6 veterinary pathologists (dual vote ground truth) or agreement by 3/6 veterinary pathologists (three vote ground truth). The $F_1$ score was used as a metric to compare the algorithmic approaches and six experts to the GT labels. We considered two annotations to be identical if the centers were within a Euclidean distance of 25\,px (equivalent to approximately the mean radius of a  mast cell).

\subsection{Co-localisation of bi-, multi-nucleated and mitotic tumor cells}
In order to better understand the pathology task of enumerating BiNC and MuNC, we correlated the density of BiNC and MuNC with mitotic figures (previously published  for these WSI \cite{03}). We used a moving window summation with an area of 2.37\,mm$^2$, as described in Auberville et al. \cite{02}, to derive the mitotic count (MC), as well as the BiNC count and the MuNC count. To compare those density metrics, we used Pearson's correlation coefficient.

\section{Results}
Our deep learning-based model yielded an $F_1$ score of 0.675 for BiNC and 0.623 for MuNC when assessed on the dataset GT of the entire 11 test WSI. Comparing the algorithm to the performance of six veterinary pathologists on regions of interest of the test WSI, the algorithm outperformed all pathologists for detection of BiNC (Tab. \ref{tab1}) and MuNC (Tab.  \ref{tab2})  when using the dataset GT definition, but severely deteriorated for multi-expert GT definitions. 
In our assessment, the expert performance varied significantly (inter-observer variability) regardless of the GT definition, whereas identification of MuNC generally had higher accuracy than identification of BiNC (Fig. \ref{fig1}). This is in contrast to the algorithmic performance which was hampered by lower numbers of labels on MuNC (compared to BiNC) for training the model. 

Co-localisation of BiNC cells with MuNC (r = 0.66) had a higher positive correlation than co-localisation of BiNC or MuNC with mitotic density (r = 0.42 and 0.29, respectively).

\begin{table}[t]
\caption{Performance ($F_1$ score) of veterinary pathologists (VP 1-6) and our deep learning-based model (DL) for detecting binucleated tumor cells compared to different definitions of ground truth (GT) labels.}
\label{tab1}
\begin{tabular*}{\textwidth}{l@{\extracolsep\fill}ccccc}
\hline
 \multirow{2}{*}{} &  \multicolumn{3}{c}{Region of interest test set} & \multirow{2}{*}{WSI test set} \\
  \cline{2-4}
  &  Three-vote GT & Dual-vote GT  & Dataset GT \\

%VP vs. DL & Majority vote GT  & Dual vote GT & Dataset GT & Dataset GT \\ 
\hline
VP 1 & 0.513 & 0.567 & 0.526 & N/A \\
VP 2 & 0.524 & 0.420 & 0.328 & N/A \\
VP 3 & 0.392 & 0.478 & 0.280 & N/A \\
VP 4 & 0.424 & 0.433 & 0.270 & N/A \\
VP 5 & 0.261 & 0.511 & 0.300 & N/A \\
VP 6 & 0.470 & 0.372 & 0.329 & N/A \\
Median VP 1-6 & 0.447 & 0.455 & 0.314 & N/A \\
\hline
DL & 0.438 & 0.424 & 0.667 & 0.675\\
\hline
\end{tabular*}
\end{table}

\begin{table}[t]
\caption{Performance ($F_1$ score) of veterinary pathologists (VP 1-6) and our deep learning-based model (DL) for detecting multinucleated tumor cells compared to different definitions of ground truth (GT) labels.}
\label{tab2}
\begin{tabular*}{\textwidth}{l@{\extracolsep\fill}ccccc}
\hline
 \multirow{2}{*}{} &  \multicolumn{3}{c}{Fields of interest test set} & \multirow{2}{*}{WSI test set} \\
  \cline{2-4}
  &  Three vote GT & Dual vote GT  & Dataset GT \\

%VP vs. DL & Majority vote GT  & Dual vote GT & Dataset GT & Dataset GT \\ 
\hline
VP 1 & 0.610 & 0.556 & 0.622 & N/A \\
VP 2 & 0.508 & 0.574 & 0.513 & N/A \\
VP 3 & 0.355 & 0.478 & 0.361 & N/A \\
VP 4 & 0.613 & 0.559 & 0.545 & N/A \\
VP 5 & 0.466 & 0.627 & 0.441 & N/A \\
VP 6 & 0.375 & 0.360 & 0.316 & N/A \\
Median VP 1-6 & 0.487 & 0.558 & 0.477 & N/A \\
\hline
DL &0.481 & 0.4 & 0.685 & 0.628 \\

\hline
\end{tabular*}
\end{table}

\section{Discussion}
With this publication we present a novel open access dataset on BiNC and MuNC in ccMCT. A first deep learning-based model was able to yield an $F_1$ score of above 0.6 for both label classes and outperformed all pathologists in object detection compared to the dataset GT. The model architecture was based upon previous publications on algorithms for mitotic figure detection \cite{02,03}. It was beyond the scope of the present study to compare performance of different algorithmic approaches, however, we encourage other research groups to use this publicly available dataset to improve state-of-the-art methods for this task.

Although pathologists are the gold standard for labeling structures in histologic sections, an object detection challenge revealed high inter-rater variability for identifying BiNC and MuNC, which is likely also reflected in label accuracy of the dataset. Therefore, a limitation of the present dataset could be that it was only created by a single pathologist. Mitotic figure datasets have usually used the consensus by at least two experts \cite{03,10}, which intents to improve label consistency. 
Obstacles to overcome when classifying BiNC and MuNC include unclear visualization of cell boarders between adjacent cells as well as differentiation from imposters with indented or lobulated nuclear shape. Furthermore, overlooking/omission of objects was a common source of error. Apart from containing additional nuclei, BiNC are often inconspicuous, which makes them difficult to recognize. Enlargement of cell size was more common in MuNC, which could explain somewhat higher sensitivity of identifying these objects. For dataset creation, we therefore decided to use an algorithmic-augmented labeling approach that was able to detect many missed candidates. As the same model was used to detect potential candidates and to evaluate performance, a certain bias towards including objects that are easy detected by the model cannot be neglected. 
In order to reduce the bias of this labeling approach we intentionally used a low detection threshold (many false positves) and all labels were reviewed by an expert . Our dataset likely includes a high degree of variability regarding tissue morphologies as entire WSI were labeled. However, additional datasets may be necessary to increase variability resulting from different staining protocols and use of different whole slide scanners.

Positive correlation of BiNC and MuNC with mitotic density suggests that endoreplication is a plausible mechanism for their formation in ccMCT. Although the number of MuNC is am important prognostic parameter of ccMCT, the number of BiNC in histologic tumor secions has not been correlated to patient outcome to date (as opposed to cytologic specimens \cite{07}). As the density of BiNC was proportional to that of MuNC and both are evidence of polyploidy, we speculate that histologic assessment of binucleation may have prognostic relevance in ccMCT. Whereas MuNC are very sparse or absent in most ccMCT cases and thus is predictive of poor outcome in only small numbers of cases \cite{04}, BiNC are more common in ccMCT and should be considered as a potential prognostic parameter in future studies. Due to the high inter-observer variability observed in this study we highlight that methods of enumeration should be better standardized and we propose that deep learning-based models may be useful to increase reproducibility and possibly accuracy for assessment of this parameter.

\subsection*{Acknowledgements} 
CAB gratefully acknowledges financial support received from the Dres. Jutta \& Georg Bruns-Stiftung f\"ur innovative Veterin\"armedizin.

\end{document}